# Unsupervised Anomaly Detection with an Enhanced Teacher for Student-Teacher Feature Pyramid Matching

Mohammad Zolfaghari
*Department of Computer Science,*
*University of Tehran*
Kish International Campus, Iran
mmzolfaghari@ut.ac.ir

Hedieh Sajedi
*Department of Computer Science,*
*School of Mathematics, Statistics and*
*Computer Science,*
*College of Science,*
*University of Tehran*
Tehran, Iran
hhsajedi@ut.ac.ir

*Abstract*— Anomaly detection or outlier is one of the challenging subjects in unsupervised learning. This paper is introduced a student-teacher framework for anomaly detection that its teacher network is enhanced for achieving high-performance metrics. For this purpose, we first pre-train the ResNet-18 network on the ImageNet and then fine-tune it on the MVTech-AD dataset. Experiment results on the image-level and pixel-level demonstrate that this idea has achieved better metrics than the previous methods. Our model, Enhanced Teacher for Student-Teacher Feature Pyramid (ET-STPM), achieved 0.971 mean accuracy on the image-level and 0.977 mean accuracy on the pixel-level for anomaly detection.

## I. Introduction

Anomaly detection has an essential role for intelligent agents to comprehend if an observed pattern is normal or abnormal. Today, automatic anomalies detection in data is very important [1], [2]. Many problems from various vision applications are anomaly detection, including construction defect intuition [2], [3], medical image analysis [4], [5] and video surveillance [6], [7]. Unlike a typical supervised classification problem, anomaly detection faces unique challenges. First, due to the nature of the problem, it isn't easy to obtain a large amount of anomalous data, either labeled or unlabeled. Second, the difference between normal and abnormal patterns is often fine-grained as defective areas might be small and subtle in high-resolution images [8]. Unsupervised or semi-supervised learning has been used for anomaly detection in previous studies [9]. Whereas the distribution of anomaly patterns is unknown in progress, models are trained to learn patterns of normal instances and determine anomalies if the test sample is not represented fine by these models [8], [10]. Generative models declare anomalies when the probability density is below a certain threshold. Alternative methods using high level learned representations have shown more effective for anomaly detection. For example, a deep one-class classifier demonstrates an effective end-to-end trained one-class classifiers parameterized by DNNs[1]. It outperforms its shallow counterparts, such as one-class SVMs[2] and reconstruction-based approaches, such as autoencoders. DNNs such as CNNs[3] and ResNets[4] extract powerful features from data [11]–[14].

## II. Related Works

In recent years, some efforts have been performed to diagnose automated anomalies. Generally, the pre-trained networks on image classification are applied to image-level detection [15], [16].

Cohen and Hoshen proposed SPADE[5] model that extracted a set of features from a pre-trained network [17]. The method is achieved acceptable results between accuracy and complexity in the anomaly identification task.

Bergmann et al. investigated a student-teacher network for unsupervised anomaly detection [18]. The difference between the output of the student network and the teacher network determined the network's final output.

Wang et al. proposed student-teacher architecture for anomaly detection called STPM[6] [10]. The STPM is made with the pre-trained network as the teacher that the knowledge is distilled into the student network with the same architecture. The learning process of the student network is performed with the distribution of anomaly-free images by matching their features. The hierarchical architecture of STPM is applied for identifying the different sizes of anomalies. The teacher and student feature pyramids are compared for prediction where the probability of anomaly occurrence is increased.

---

[1] Deep Neural Networks(DNNs)
[2] Support Vector Machines(SVMs)
[3] Convolutional Neural Networks(CNNs)
[4] Residual Networks(ResNets)
[5] Semantic Pyramid Anomaly Detection(SPADE)
[6] Student-Teacher Feature Pyramid Matching(STPM)

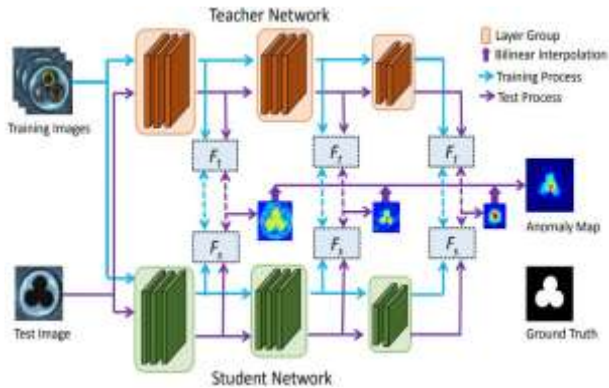

Fig. 1. The STPM model [10]

## III. METHOD

This paper is presented an approach that the teacher network is enhanced. Some anomalous data are used in the training step to fine-tune the teacher on the task-specific dataset. The fine-tuning step of the teacher network has been done before the anomaly detection step. So, the pre-trained ResNet-18 network is employed, and it has been fine-tuned on the MVTech-AD dataset[7] [19] normal and some abnormal images. Sample normal and abnormal samples of the MVTech-AD dataset are presented in Fig. 2. The ET-STPM achieves better teacher performance on extracting the features of anomalous images that the student didn't see. So, this makes a larger difference in feature pyramids from the teacher and student networks that are larger than the STPM differences. Larger differences can make anomaly detection more reliable and efficient.

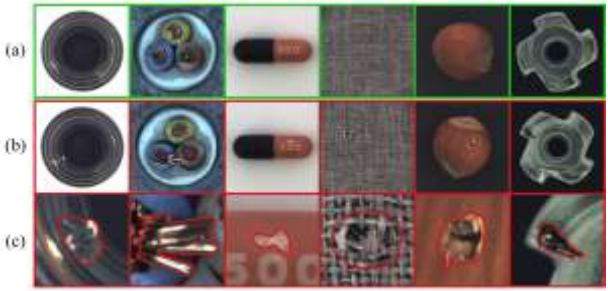

Fig. 2. Some images from MVTech-AD dataset: (a) Normal (b) Abnormal (c) The damaged regions of the abnormal images.

In the following, the ET-STPM architecture, training and test processes are explained:

### A. Framework

The ET-STPM model consists of two parts: fine-tune the base network and student-teacher framework.

### B. Fine-Tuning

The ResNet-18 is pre-trained on the ImageNet dataset [20]. Although this dataset is completely reachable but we fine-tune the network weights on task-specific images. The specific images are shown by defining a classification task want to on the dataset classes. The last Fully Connected layer of the ResNet-18 classifier has 512 input neurons and 1000 output neurons, but the MVTech-AD dataset has 15 classes. So, we changed this Fully

---
[7] MVTec Anomaly Detection dataset(MVTech-AD)

---

Connected layer with a 15 output neurons layer. The weights of this layer are random. Therefore, the weights of all layers, except this new layer, are frozen, and the Fully Connected layer of a classifier scheme with one epoch has been trained. The initial random weights of this layer can cause large weight changes in feature selection layers. So, we can unfreeze the entire network and train all layers for two epochs to achieve the best performance metrics. Experiments on a number of epochs are showed that a little over-fitting in this step and reaching the highest accuracy can improve model performance at anomaly detection task because a little over-fitting on this dataset makes the extracted features of the network more related to the task instead of global ImageNet features.

### C. Student-Teacher Architecture

ResNet-18 is pre-trained on ImageNet and builds the teacher network. The student network has a similar teacher network structure for reducing information loss. Feature extraction is done with the bottom layers of the teacher network that helps the student network learning. In addition, this structure detects anomalies of different sizes.

### D. Training Process

Fine-tuning and the student-teacher are two steps of the training process. The first step is performed by a classification scheme on the MVTech-AD dataset classes with a Cross-Entropy loss function. We define the loss and use it in the second step. So, L1 norm and cosine similarity distance are used to feature vectors at position (x, y) of the feature maps.

### E. Test Process

The test process is performed by using some of the images in fine-tuning step, but the classification task is easy. So, the train and test accuracy are both 100%. It is worth mention that the accuracy of fine-tuning step is not important because this classification task is only used for tuning the feature selection layers more task-specific and can change by any other schemes e.g., unsupervised schemes.

## IV. EXPERIMENT

We utilize the first three blocks of ResNet-18 as the pyramid feature extraction. ResNet-18 pre-trained on ImageNet is sent the corresponding values to the teacher network, but the student network is initialized randomly. The training parameters of the model are set according to Table I.

TABLE I. THE MODEL PARAMETERS

| Optimizer | Momentum | Batch Size | Weight Decay | Learning Rate | Epochs |
|---|---|---|---|---|---|
| SGD [a] | 0.9 | 32 | $10^{-4}$ | 0.4 | 100 |

[a.] Stochastic Gradient Descent (SGD)

Python programming language is applied to implement the ET-STPM on the MVTec-AD dataset to consider both image-level and pixel-level anomaly detection. The image-level and pixel-level anomaly detection results are shown in Tables II and III, respectively.

TABLE II. COMPARISON RESULTS FOR IMAGE-LEVEL ANOMALY DETECTION

| GANomaly [21] | l2-AE [22] | ITAE [23] | SPADE | STPM | ET-STPM |
|---|---|---|---|---|---|
| 0.762 | 0.754 | 0.839 | 0.855 | 0.955 | **0.971** |

The performance metric of Table III is calculated by mean AUC-ROC[8] through 15 classes. The ET-STPM is attained 0.971 average accuracy on the image-level and 0.977 average accuracy on the pixel-level anomaly detection. Therefore, we improve average performance metrics 1.6% in the image-level and 0.7% in the pixel-level anomaly detection compared with the second-best model (STPM).

TABLE III. COMPARISON RESULTS FOR PIXEL-LEVEL ANOMALY DETECTION

| Category | | AnoGAN | CNN-Dict | SPADE | STPM | ET-STPM |
|---|---|---|---|---|---|---|
| *Texture* | Carpet | 0.54 | 0.72 | 0.975 | 0.988 | **0.997** |
| | Grid | 0.58 | 0.59 | 0.937 | 0.990 | **0.993** |
| | Leather | 0.64 | 0.87 | 0.976 | 0.993 | **0.995** |
| | Tile | 0.50 | 0.93 | 0.874 | 0.974 | **0.988** |
| | Wood | 0.62 | 0.91 | 0.885 | 0.972 | **0.980** |
| *Objects* | Bottle | 0.86 | 0.78 | 0.984 | **0.988** | 0.981 |
| | Cable | 0.78 | 0.79 | **0.972** | 0.955 | 0.949 |
| | Capsule | 0.84 | 0.84 | **0.990** | 0.983 | 0.988 |
| | Hazelnut | 0.87 | 0.72 | **0.991** | 0.985 | 0.990 |
| | Metal nut | 0.76 | 0.82 | 0.981 | 0.976 | **0.985** |
| | Pill | 0.87 | 0.68 | 0.965 | 0.978 | **0.980** |
| | Screw | 0.80 | 0.87 | **0.989** | 0.983 | 0.985 |
| | Toothbrush | 0.93 | 0.90 | 0.979 | **0.989** | 0.980 |
| | Transistor | 0.86 | 0.66 | **0.984** | 0.825 | 0.869 |
| | Zipper | 0.78 | 0.76 | 0.965 | 0.985 | **0.997** |
| ***Mean*** | | 0.74 | 0.78 | 0.965 | 0.970 | **0.977** |

Visual results of the ET-STPM on three damaged images from the MVTec-AD dataset are shown in Fig. 3. Ground Truth regions are shown with red color. Columns from left to right indicate damaged input image, anomaly maps of the three bottom convolutional blocks of ResNet-18 and the final anomaly maps respectively.

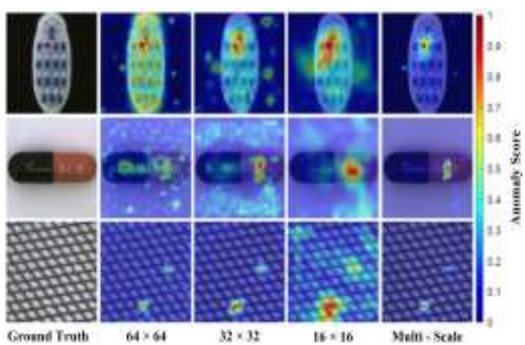

Fig. 3. Visual results of the three damaged images.

## V. CONCLUSION

We developed the teacher network of the STPM model and called it ET-STPM. Using the Pre-trained ResNet-18 network for the teacher network of the STPM method is created a more enhanced student-teacher architecture for anomaly detection. So, we executed the pre-trained ResNet-18 network on the ImageNet dataset and used it in the teacher network of the ETSTPM. The ET-STPM is executed on the MVTec-AD dataset and get better results contrasted to the previous models on both image and pixel levels.

---

[8] Area Under the Curve-Receiver Operating Characteristics (AUC-ROC)